\documentclass{article}
\usepackage{ijcai24}

\usepackage{times}
\usepackage{soul}
\usepackage{url}
\usepackage[hidelinks]{hyperref}
\usepackage[utf8]{inputenc}
\usepackage[small]{caption}
\usepackage{graphicx}
\usepackage{amsmath}
\usepackage{amsthm}
\usepackage{booktabs}
\usepackage{algorithm}
\usepackage{algorithmic}
\usepackage[switch]{lineno}
\usepackage{enumitem}
\usepackage{xcolor}
\usepackage{multirow}
\usepackage{caption}

\title{Assessing Empathy in Large Language Models with Real-World Physician-Patient Interactions}

\author{
Man Luo$^{1,2}$
\and
Christopher J. Warren$^1$\and
Lu Cheng$^{3}$\and
Haidar M. Abdul-Muhsin$^1$\\
Imon Banerjee$^1$
\affiliations
$^1$Mayo Clinic\and
$^2$ Intel Lab\and
$^3$ University of Illinois Chicago\\
\emails
Man.Luo@intel.com\\
\{warren.christopher,abdul-muhsin.haidar,banerjee.imon\}@mayo.edu\\
lucheng@uic.edu
}

\begin{document}

\maketitle
\begin{abstract}
The integration of Large Language Models (LLMs) into the healthcare domain has the potential to significantly enhance patient care and support through the development of empathetic, patient-facing chatbots. 
This study investigates an intriguing question \textit{Can ChatGPT respond with a greater degree of empathy than those typically offered by physicians?} 
To answer this question, we collect a de-identified dataset of patient messages and physician responses from {Mayo Clinic} and generate alternative replies using ChatGPT. Our analyses incorporate novel empathy ranking evaluation (EMRank) involving both automated metrics and human assessments to gauge the empathy level of responses. Our findings indicate that LLM-powered chatbots have the potential to surpass human physicians in delivering empathetic communication, suggesting a promising avenue for enhancing patient care and reducing professional burnout. The study not only highlights the importance of empathy in patient interactions but also proposes a set of effective automatic empathy ranking metrics, paving the way for the broader adoption of LLMs in healthcare.
\end{abstract}

\section{Introduction}
\begin{figure*}
    \centering
    \includegraphics[width=\linewidth]{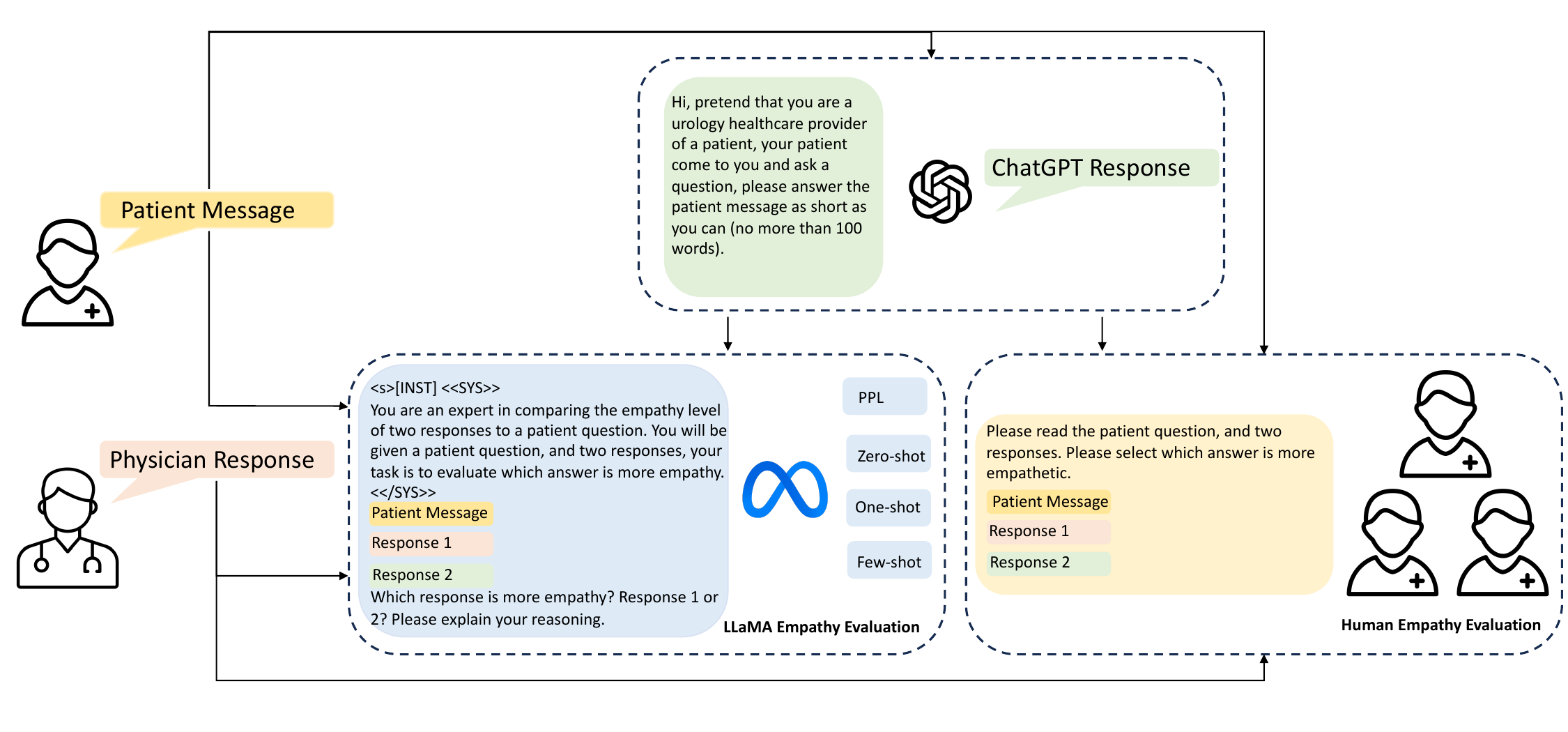}
    \caption{Given a patient message, we prompt ChatGPT for a response. We restrict the length of the ChatGPT response to mimic the statistics of the physician's response (See Table \ref{tab:dataset_statistic}). Both ChatGPT's and a physician's responses are then evaluated using a multi-dimension LLM-EMRank metric for automatic ranking (LLaMA Empathy Evaluation). In addition. we also conduct a human empathy evaluation to ensure a thorough and rigorous assessment.}
    \label{fig:eval}
\end{figure*}


Large Language Models (LLMs) have exhibited exceptional performance in clinical and biomedical areas~\cite{lee2020biobert,luo2022biogpt}. They have been effectively used for extracting information from clinical notes~\cite{parmar2022boxbart,chiang2023large}, providing differential diagnoses based on patient symptoms~\cite{macherla2023mddial}, retrieving biomedical literature information~\cite{luo2022improving},
and assisting data labelling~\cite{he2024if}.
underscoring their flexibility and vital role in enhancing healthcare services and research advancements. 
A notable achievement, such as with models like ChatGPT~\cite{ray2023chatgpt}, is the development of patient-facing chatbots built on the foundation of LLMs~\cite{dave2023chatgpt}. These chatbots aim to deliver instant answers to patient questions, significantly reducing the waiting period for doctor's advice and helping to decrease the burden on healthcare professionals~\cite{wilson2022development}. Beyond meeting immediate patient needs, chatbots that respond with empathy can greatly boost patient engagement, a key factor in improving patient quality of life (QoL). An empathetic chatbot is more effective in maintaining continuous dialogue with patients, creating a nurturing space even within the digital realm~\cite{inkster2018empathy}.

While the creation of empathetic dialogues in general contexts has been explored~\cite{tahir2023artificial}, a comprehensive examination of chatbots' empathetic abilities when interacting with real patients and comparing their responses to those of actual physicians remains limited~\cite{altamimi2023artificial}. This study \textbf{aims to} answer an intriguing question, \textit{Can ChatGPT respond with a greater degree of empathy than those typically offered by physicians?} Particularly, we collect a real-world dataset comprising patient messages and the corresponding responses from physicians at {Mayo Clinic}. We manually de-identify patients' and physicians' information to protect their privacy. Subsequently, we utilize the ChatGPT API\footnote{https://platform.openai.com/docs/overview}, a leading-edge LLM, to craft responses to these de-identified patient messages, following detailed instructions (refer to the green box in Figure~\ref{fig:eval}).

The next step of our research involves assessing the level of empathy in responses from both physicians and ChatGPT. Previous studies have primarily focused on detecting empathy through models trained on specific datasets, a technique that often lacks generalizability to different domains without domain-specific fine-tuning. 
Drawing inspiration from the emerging ability of LLMs~\cite{wei2022emergent}, especially the in-context learning, where an LLM can perform an unseen task with instruction and few examples of the task, we investigate the effectiveness of in-context learning for empathy ranking in healthcare domain. 
We introduce multiple metrics using LLaMA~\cite{touvron2023LLaMA}, named LLaMA-EMRank. Specifically, we employ LLaMA zero-shot, one-shot, and few-shot learning capabilities, as well as an ensemble of these methods.
In a zero-shot scenario, we prompt LLaMA to execute the EMRank task through domain-specific instructions (refer to the blue box in Figure~\ref{fig:eval}).
For one-shot and few-shot scenarios, we collect in-context learning (ICL) examples from patients (\S\ref{sec:llama-emrank}). 
Compared with prior metrics~\cite{lee2022improving,kim2021perspective,lee2022does}, 
LLaMA-EMRank eliminating the need of fine-tuning an evaluation model and the framework is generalizable to other domains (i.e. by changing the few-shot ICL examples and domain specific instruction). 

\textcolor{black}{
We strategically choose a strong model (i.e. ChatGPT) to be the response generation model and a smaller model (i.e. LLaMA) to be the empathy evaluation model for two reasons. 
First, empathetic response generation to a patient message task is a complex task,   necessitating not just an understanding of empathy but also a comprehensive and accurate grasp of the relevant domain knowledge, areas where ChatGPT excels as a leading model. 
Second, given that ChatGPT is the response generation model, to ensure independence in the assessment process, minimizing any potential bias that might arise if the same model (ChatGPT) were used for both generating and evaluating responses, choose a different LLM that preserves robust linguistic understanding ability. The concept of weak-to-strong generalization has been explored~\cite{burns2023weak}.
Moreover, our experiments show that the LLaMA-EMRank metrics reach a desired agreement with humans evaluation, indicating its validity of ranking the empathy degree of two responses  (\S\ref{sec:llama-experiments}).
}

Overall, our investigation aims to measure the empathetic quality of LLM-powered chatbot responses in comparison to human physicians and to examine how well these models' outputs align with human judgments. 
Through this work, we aspire to reveal the capabilities of LLMs in fostering patient care and support, representing a significant advancement in integrating artificial intelligence into healthcare.
We summarize the three contributions of this work: 

\begin{itemize}[noitemsep,leftmargin=*] 
    \item Our analysis is based on real-patient data, ensuring that the findings and conclusions are directly relevant to real-world healthcare scenarios.
    \item We compares the empathetic responses of ChatGPT with those of physicians using four innovative automatic metrics: LLaMA-EMRank.
    \item we incorporate patient evaluations to assess the concordance between LLaMA-EMRank metrics and human perceptions, demonstrating the metrics' reliability.
    \item Our findings suggest potential for LLM-powered chatbots to surpass physicians in delivering empathetic responses, thereby enhancing patient interaction.
\end{itemize}

\section{Related Work}
\begin{table*}[t]
\centering
 \resizebox{\linewidth}{!}{
\begin{tabular}{c|c|c|c|c}
    \toprule
    \textbf{Dataset} & \textbf{Domain} & \textbf{Source} & \textbf{Curating Method} & \textbf{Evaluation Metric} \\
     \toprule
    EmpatheticDialogues~\citeyear{rashkin2019towards} & Daily & AMT Conversation & Crowd-sourcing & PPL , BLUE, UE-score, EPITOME,Diff-EPITOME\\
    EMPATHETICPERSONAS~\citeyear{rashkin2019towards} &  Mental Health & AMT Conversation & Crowd-sourcing & PPL, Accuracy \\
    IEMPATHIZE~\citeyear{hosseini2021takes} & Oncology & Online Cancer Platform & Graduate Student Annotation & F1, Precision, Recall\\ 
    EDOS~\citeyear{welivita2021large} & Daily & Movie Dialogue & semi-automated & F1, Accuracy\\
    EPITOME~\citeyear{sharma2020computational}& Mental Health & Online support platforms & Crowd Sourcing & F1, Accuracy\\
    \midrule
    Ours & Oncology & Hospital & Real-patient & PPL, zero-shot, one-shot, few-shots\\
    \bottomrule
    \end{tabular}
    }
\caption{Comparing with Existing Datasets and Metrics for Empathy Evaluation.}
\label{tab:related_work}
\end{table*}

\paragraph{Empathy Dataset} 
Several dialogue datasets have been annotated with emotion labels, demonstrating the diversity in approaches and applications within this research area~\cite{busso2008iemocap,chatterjee2019understanding}. Notably, \cite{welivita2020taxonomy} conducted an in-depth analysis of the EmpatheticDialogues dataset~\cite{rashkin2019towards}, identifying eight empathetic response intents, such as encouragement and consolation, alongside 32 fine-grained emotions that range from positive to negative. The process of labeling these datasets varies, involving either human annotators or semi-automatic methods, which occasionally results in ambiguous emotion categorizations~\cite{chatterjee2019understanding}.
\cite{rashkin2019towards} employed Amazon Mechanical Turk (AMT) workers to gather a dataset enriched with textual expressions spanning four emotional contexts: sadness, anger, anxiety/fear, and happiness/contentment. This dataset also features empathetic rewritings of 45 foundational utterances. In a similar setting, EPITOME~\cite{sharma2020computational} aggregates conversations from online mental support platforms, with AMT workers tasked to detect three key communication mechanisms: emotional reactions, interpretations, and explorations.
Moreover, \cite{hosseini2021takes} introduced the IEMPATHIZE dataset, which comprises sentences sourced from an online platform for cancer patients (specifically those with lung and breast cancer). This dataset was annotated by two graduate students who classified each sentence into one of three categories: seeking empathy, providing empathy, or none, highlighting the nuanced understanding of empathetic communication within specific contexts.
The EDOS dataset is curated through a semi-automatic approach by~\cite{welivita2021large}. It begins with a selection of 9,000 movie subtitles from the OpenSubtitles corpus, which are initially annotated by humans to match the emotion labels used in EmpatheticDialogues. These annotated subtitles are then used to train a BERT-based emotion classification model, which in turn annotates additional movie subtitles, culminating in a dataset comprising over one million entries.

Our research uniquely contributes by analyzing real conversations between patients and physicians, with patients comparing chatbot to physician responses, offering insights closest to real-world healthcare scenarios. Unlike prior studies focused on emotion detection or dialogue generation, our work centers on evaluating chatbot versus physician interactions, aligning with the trend of using LLMs as chatbots without domain-specific tuning. This direct comparison approach not only differentiates our dataset but also highlights the practical application of LLMs in healthcare, bridging a significant gap in existing research.

\paragraph{Empathy Evaluation}
In the realm of emotion detection tasks, accuracy and F1 scores are commonly used to assess system performance, as indicated in Table \ref{tab:related_work}. For evaluating response fluency, BLEU and perplexity (PPL) metrics are employed. These methods operate independently of fine-tuned models, contrasting with metrics that necessitate model fine-tuning to gauge empathy levels. 
\cite{lee2022improving} introduced the UE-score (utterance entailment), utilizing a model trained on the SNLI dataset to determine if a hypothesis is entailed by a premise. \cite{kim2021perspective} developed an automatic scoring mechanism based on the EPITOME framework~\cite{sharma2020computational}, employing three RoBERTa models~\cite{liu2019roberta} to predict emotional reactions, interpretations, and explorations, with the final empathy score of a sentence being the average of these three scores. Based on EPITOME, ~\cite{lee2022does} crafted the Diff-EPITOME score to compare the empathy in responses from humans and chatbots, where a lower score signifies a more human-like response. However, this assumes that humans always express empathy, which may not hold true under all circumstances. Another limitation of  EPITOME framework's is that the three categories might not work in all circumstances. For instance, \cite{october2018characteristics} outlined five empathy evaluation aspects in patient-physician interactions that differ from the EPITOME framework. Despite the reliance on automatic metrics, human evaluations play a crucial role in validating whether responses are empathetic or if automatic metrics align with human judgment. \cite{li2019acute} introduced ACUTE-EVAL for comparing responses in dialogue contexts, using Pearson's r score to assess alignment between human and automatic evaluations. \cite{jamainternmed2023} compared the empathy degree of a model generated answer and physician's answer by human annotation. 

Our study introduces unique metrics for ranking empathy levels between responses without depending on fine-tuning, setting our approach apart from previous methodologies.

\section{Data Collection}

\paragraph{Source of the Data} 
Our dataset comprises messages collected from a patient portal at {Mayo Clinic}, which is used for answering patients' questions. We collect messages from  individuals diagnosed with prostate cancer. This dataset was compiled by extracting the conversations of men who underwent radical prostatectomy between December 2018 and October 2023. 
We gather messages under the category of ``Patient advice request" and randomly select messages from the pool of available conversations. 

\paragraph{De-indefication Process} The de-identification process was meticulously carried out by a team of three two medical students and one physician who is a post graduate year 3 urology resident, with the primary goal of ensuring the privacy and confidentiality of individuals mentioned in the dataset collected from the internal platform. 
This process involved a detailed review of each patient message and physician response to identify and remove or anonymize any personally identifiable information (PII) including patient names, physician names, dates, phone numbers, and addresses.

\paragraph{Statistic}

Table~\ref{tab:dataset_statistic} shows the statistics of the dataset. Notably, ChatGPT's responses tend to be more lengthy on average than those of human counterparts, yet they align well with our specified length limit of 100 words. This observation suggests ChatGPT's effective management of response length.
Examples of patient messages and physician responses can be found in Figure~\ref{fig:one_incontext_example} and Figure~\ref{fig:three-incontext-examples} in the Appendix.

\begin{table}[t]
\centering
\begin{tabular}{c|c}
    \toprule
    \textbf{Statistic} & \textbf{Value} \\
     \toprule
    Total Number of Questions & 491 \\
    Average Length of Question & 94  \\
    Average Length of Physician Response  & 78  \\
    Average Length of ChatGPT Response  & 104  \\
    \bottomrule
    \end{tabular}
\caption{Empathy Evaluation Dataset Statistics: Analyzing Messages from Real Prostate Cancer Patients, Responses from Physicians, and ChatGPT's Replies.}
\label{tab:dataset_statistic}
\end{table}

\section{Chatbot Response}
\label{sec:chatbot_response}

\paragraph{ChatGPT}  is the state-of-the-art language model developed by OpenAI~\cite{achiam2023gpt}. Its adeptness at mimicking human conversation allows it to cover a broad spectrum of topics, offering explanations, answers, and creative outputs.
The model has notably captured the interest of the clinical sector, where its potential to improve patient care, medical education, and operational efficiency is considerable. Our research focuses on a particularly intriguing question: \textit{Can ChatGPT respond with a greater degree of empathy than those typically offered by physicians?}
To explore this, we leverage ChatGPT's instructional learning capabilities, employing specific prompts to elicit responses from the model, as detailed in the subsequent section. 
Our experiments are based on the API of ChatGPT, gpt-4-1106-preview version which is trained on the knowledge up to April 2023. We use the ChatGPT inference length of 4,096, which is the maximum number of tokens (words, punctuation, and spaces) that ChatGPT can process or generate in a single inference. 

\paragraph{Prompt} 

The specific prompt we used in our study is illustrated in the green box of Figure~\ref{fig:eval}. We instruct ChatGPT to function as a urology expert, considering our dataset comprises questions from prostate cancer patients. Additionally, we guide ChatGPT to respond with empathy and limited the response length to 100 words. Initially, we did not impose a word limit, but we observed that ChatGPT often produced lengthy responses. To ensure a fair comparison with the responses from physicians, we decided to implement a word count restriction. This decision was made after comparing the average length of physician responses with ChatGPT responses, as presented in Table~\ref{tab:dataset_statistic}.
\section{Evaluation}

Given that existing evaluation metrics primarily focus on empathy detection tasks, we explore LLM's capability for automatic empathy ranking evaluation. To this end, we propose multiple empathy ranking metric that consists of several new empathetic factors described by LLMs. 
For the analysis of ChatGPT's responses, we employ a distinct language model for comparison with those of physicians. We select the LLaMA model~\cite{touvron2023LLaMA}  for this purpose, as it stands out as one of the best open-source models, having been extensively researched and applied in various studies.
Specifically, we make use of the LLaMA model's in-context learning capabilities to assess empathy, and termed this family metrics as LLaMA-EMRank. 
Additionally, we complement this approach with a human study to gauge how well each automated method aligns with human judgment.

\subsection{LLaMA-EMRank: Automatic Empathy  Ranking Evaluation}
\label{sec:llama-emrank}

\paragraph{Zero-shot Evaluation}
LLMs have demonstrated a remarkable capacity for zero-shot learning without the need for task-specific fine-tuning~\cite{brown2020language}. This capability is particularly valuable as it eliminates the extensive computational and data preparation costs associated with fine-tuning models for specific downstream tasks. We employ a language model to determine the more empathetic response between two options: one from a physician and the other from the chatbot, as a reply to a patient's question. This involves feeding the model both responses and asking it to identify which one exhibits greater empathy. The blue box in Figure~\ref{fig:eval} showcases the inputs given to the model. 

\paragraph{Few-shot Evaluation} 

\begin{figure}[t]
    \centering
\includegraphics[width=0.95\linewidth]{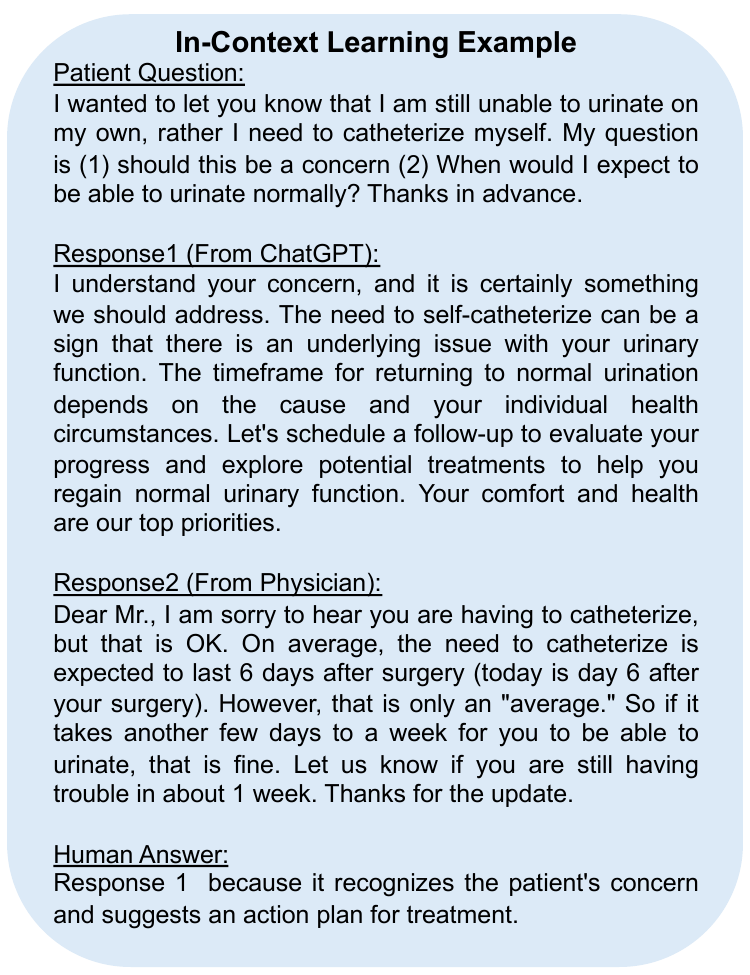}
    \caption{In Context Learning Example: Patient is given two responses and assesses which response is more empathetic and provides the justification. Note that patients do not know which response is from ChatGPT or the physician when evaluating empathy. }
    \label{fig:one_incontext_example}
\end{figure}

LLMs also excel in few-shot learning~\cite{luo2024context}, where they leverage a small number of examples to adapt to new tasks. This few-shot capability significantly enhances the models' flexibility and efficiency, allowing them to perform specialized tasks by understanding the context from a few demonstration examples.
In applying this few-shot learning to evaluate empathy in responses, we start by collecting examples by presenting a patient's question alongside two responses, who then determine which response is more empathetic and explain why. After collecting three examples, we input them into an LLM. After the three examples, we then ask LLaMA to rate a new message from the patient and two responses from the physician and ChatGPT which response is more empathetic. Figure~\ref{fig:one_incontext_example} depicts one demonstration example provided to LLaMA and the other two examples are given in the Appendix.

\paragraph{One-shot Evaluation} 
One-shot learning is a special case of few-shot learning where only one example is given to the model. 
Compared to few-shot learning, one-shot learning is more inference efficient since the input length to LLM is shorter~\cite{luo2023dr}.
Here, rather than randomly choosing one example from the three examples we collected, we used each one and ran three times inference. And then we take a majority vote out of three as the final answer. 

\paragraph{Ensemble Evaluation}
As we will analyze in the later section (\S\ref{sec:dissect}) that each of the previous methods have strengths and weaknesses of each approach, instead of depending on a singular method, we also introduce an ensemble metric by taking the majority consensus from zero-shot, one-shot, and few-shot assessments. 

\paragraph{Language Model Perplexity (PPL)} 

PPL is a key metric for assessing language model performance, especially in text generation tasks like dialogue systems~\cite{liu2016not}, where lower perplexity signifies more fluent, coherent, and realistic outputs. 
This metric also has applications in cognitive science to understand human language processing~\cite{frankenberg2019perplexity,carrasco2022fingerprints}.
In our research, we primarily employ Perplexity (PPL) as a measure of fluency. Additionally, driven by interest, we investigate the correlation between perplexity and empathy in language model responses within the experimental section.
We adapt PPL to our study as follows:

\begin{equation}
\text{PPL}(A \mid Q) = \sqrt[N]{\prod_{i=1}^{N}\frac{1}{P(a_i | Q, a_1, a_2, \ldots, a_{i-1})}}, 
\end{equation}
where $Q$ is a patient question, $A$ is an answer from either chatbot or a real physician, and $a_i$ is the $i^{th}$ token in the answer $A$. We compute the PPL scores for the physician's answer and the chatbot's answer and the one with lower PPL is considered as more empathetic.  

\subsection{Patient Evaluation}


We further carried out a human evaluation study that included three male patients with prostate cancer who had undergone radical prostatectomy within the same timeframe as our collected patient message dataset. For data collection, we used Google Forms to gather their annotations\footnote{https://www.google.com/forms/about/}. To avoid bias, we randomized the sequence of responses from both the physician and the chatbot for every question presented. A snapshot of the questionnaire is available in Figure~\ref{fig:questionnaire} in the Appendix.
Our study evaluated 70 questions, with each one being assessed by the three participating patients. To address potential subjectivity in the evaluations, we established a consensus for each response based on a majority vote. 

\section{Results and Analysis}

\subsection{Human Evaluation Result} 
We calculated the inter-annotator agreement rate using Fleiss' kappa score and discovered a negative score (i.e. -0.15), highlighting the subjective nature of feeling empathetic. 
To overcome this challenge, we select the option with the majority of votes as the more empathetic choice for each patient message. 
In Table~\ref{tab:results}, the human evaluation revealed that ChatGPT's responses are more empathetic than those of humans in 72.85\% of cases. 
This indicates that ChatGPT is perceived to exhibit more empathy than humans, highlighting its potential application in the healthcare field by providing patients with more mental support.

\subsection{LLaMA-EMRank Evaluation Result}
\label{sec:llama-experiments}

\begin{figure}[t]
    \centering    \includegraphics[width=0.9\linewidth]{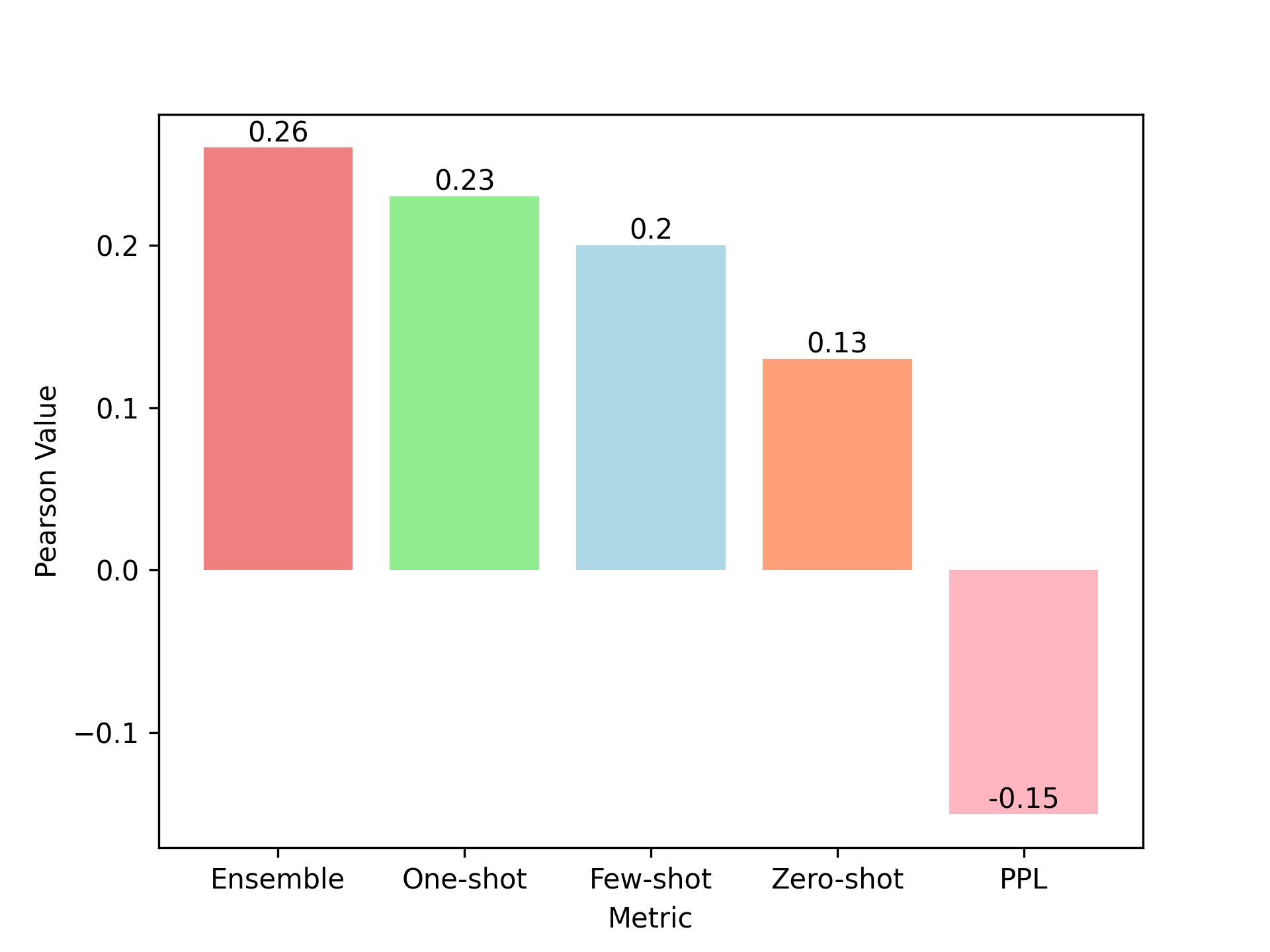}
    \caption{The Pearson values between Automatic Metric and Human Judgement.}
    \label{fig:pearson}
\end{figure}

\begin{table*}[t!]
\centering
\begin{tabular}{c|c|c|c|c|c|c}
    \toprule
    \multirow{3}{*}{\textbf{Evaluation}} &  \multicolumn{5}{c|}{\textbf{Empathy}} & \multicolumn{1}{c}{\textbf{Fluency}} \\
    \cmidrule(lr){2-6} \cmidrule(lr){7-7} 
    & \multicolumn{4}{c|}{\textbf{LLaMA-EMRank}} & \multirow{2}{*}{\textbf{Human Evaluation}} & \multirow{2}{*}{\textbf{PPL}}\\
    \cmidrule(lr){2-5} 
    &  {\textbf{Zero-shot}}  & {\textbf{One-shot}}  & {\textbf{Few-shots}}  & {\textbf{Ensemble}} &  &  \\
     \toprule
    ChatGPT  &  92.41\% & 79.09\% & 91.6\%  & 91.85\% & 72.85\% & 92.0\% \\
    Physician & 7.59\%  & 20.91\%  & 8.40\% & 8.15\% & 27.15\% & 8.00\% \\
    \bottomrule
    \end{tabular}

\caption{Comparison of ChatGPT and Physician Responses Using LLaMA-EMRank Metrics and Human Evaluation.}
\label{tab:results}
\end{table*}
We report results for automatic evaluation using LLaMA-EMRank in Table~\ref{tab:results}. We observe that across all automatic evaluation metrics, ChatGPT demonstrates a much greater empathy level than the physician. For example, for zero-shot evaluation in the third column, 92.41\% ChatGPT responses are more empathetic than Physicians. 
Of particular interest is that one-shot evaluation is the closest to the ratio of human judgment. 
However, it's noteworthy that ChatGPT's performance is considered higher on automatic metrics compared to human evaluations. 
We further investigate the correlation between human judgment and automatic metrics through Pearson's correlation coefficient, with a 95\% confidence interval.
Figure~\ref{fig:pearson} shows the results and all LLaMA-EMRank achieve positive Pearson'r value, suggesting that the introduced EMRank metrics serve as reliable indicators of the level of empathy. 
Among all metrics, ensemble metric aligns most closely with human judgment, followed by the order of one-shot voting, few-shot voting, and zero-shot. This indicates the effectiveness of ensemble method. 

Besides empathy, we also use PPL to measure the fluency of the response, and we see that in 92\% cases, ChatGPT responses show lower PPL values than the Physician's response, indicating that ChatGPT's responses are fluent. We also investigate the correlation between PPL scores and empathy. 
The negative Pearson's result in Figure~\ref{fig:pearson} shows that PPL is not a strong indicator of empathy.

\subsection{A Closer Look at LLaMA-EMRank} 
\label{sec:dissect}
Previous findings indicate that each LLaMA-EMRank metric aligns with human judgment; however, we have encountered certain limitations with each method. In this section, we will discuss the advantages and disadvantages of each LLaMA-EMRank metric and outline the strategies we have adopted to enhance their reliability.

\paragraph{Zero shot}
In this setting, we observe that for some cases, LLaMA does not generate which responses are better, rather it says ``Please let me know how I can assist you''.
To mitigate this, we have tried different prompts to identify the most effective prompt. 
Given the limitless possibilities in prompt design, we conducted several iterations of prompt exploration before settling on the most effective one.
Still, the model fails to explicitly identify which response demonstrates greater empathy. We speculate that this ambiguity arises from the nuanced complexity of the responses, which may be nearly indistinguishable in terms of empathetic content.
Another significant challenge involves the post-processing of the model's output, as it has the capability to generate a wide range of sentences. 
We attempted to constrain the output's format through the prompt, yet the model did not consistently adhere to these guidelines. 
Consequently, we manually review the LLaMA output to discover patterns and employ regular expression tools to systematically extract the evaluations from the model's output. The patterns are given in Appendix. 

\paragraph{Few shots} 
\begin{table}[t]
\centering
 \resizebox{\linewidth}{!}{
\begin{tabular}{c|c|c|c|c}
    \toprule
    \textbf{Evaluation} & {\textbf{Order 1}} &  {\textbf{Order 2}}  & {\textbf{Order 3}} & {\textbf{Majority Vote}} \\
     \toprule
    ChatGPT  & 89.0\%  & 94.91\% &  90.22\% & 91.60\%\\
    Physician & 11.0\% & 5.09\% &  9.78\% & 8.40\% \\
    \midrule 
    Pearson & 0.16 & 0.18 &\textbf{ 0.26} & 0.20 \\
    \bottomrule
    \end{tabular}
    }
\caption{Performance of Few-Shot Evaluation Across Three Variations of In-Context Examples Ordering.}
\label{tab:fewshot_results}
\end{table}
The post-processing of the model's answer is simplified due to its adherence to the format provided by in-context examples. Unlike the Zero-shot scenarios, the explanations are significantly briefer. The performance in Few-shot settings is influenced by the sequence in which examples are presented to the model. We shuffle the order of the responses three times, we observed variability in performance, as shown in Table~\ref{tab:fewshot_results}. 
The concordance between Few-shot outcomes and human evaluation also varies.
Ultimately, we consolidate the three responses and determine the final assessment based on the majority opinion.

\paragraph{One Shot}
\begin{table}[t]
\centering
 \resizebox{\linewidth}{!}{
\begin{tabular}{c|c|c|c|c}
    \toprule
    \textbf{Evaluation} & {\textbf{Example 1}} &  {\textbf{Example 2}}  & {\textbf{Example 3}}  & {\textbf{Majority Vote}}\\
     \toprule
    ChatGPT  & 71.89\% & 89.61\% & 75.76\% & 79.09\%  \\
    Physician & 28.11\% & 10.39\%  & 24.24\%  & 20.91\% \\
    \midrule
    Pearson & 0.13 & 0.20 & 0.20  & \textbf{0.23}\\
    \bottomrule
    \end{tabular}
    }
\caption{Performance of One-Shot Evaluation Performance with Different One-shot In-Context Examples.}
\label{tab:oneshot_results}
\end{table}
The post-processing of the one-shot evaluation is also much more straightforward compared to the zero-shot as the model generates the answer in the same format as the one-shot in-context example. 
Nevertheless, the outcomes of one-shot evaluations fluctuate significantly with changes in the example used, as evidenced by the large variance detailed in Table~\ref{tab:oneshot_results}. The degree of alignment between various one-shot examples and human judgment also varies. Notably, employing a majority vote approach enhances the concordance between human assessments and one-shot evaluations.
Therefore, it underscores the importance of conducting multiple one-shot evaluations and adopting the majority vote to arrive at the final judgment.



\subsection{Qualitative Analysis}

\begin{figure}[t!]
    \centering
    \includegraphics[width=0.99\linewidth]{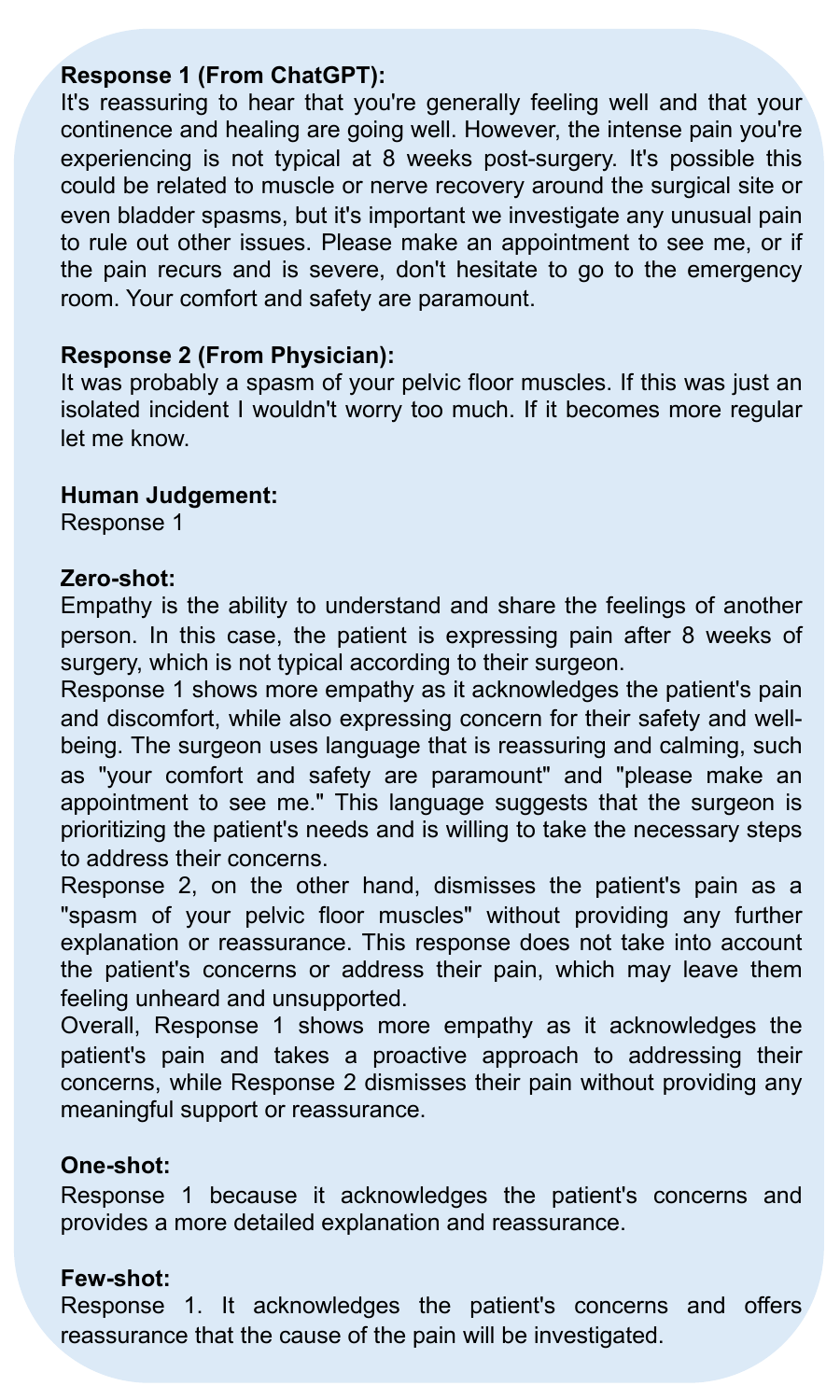}
    \caption{Given the responses from ChatGPT and Physician, human, and all other automatic metrics rate ChatGPT's response as being more empathetic. Note that patients do not know which response is from ChatGPT or the physician.}
    \label{fig:model_is_better}
    \vspace{-6pt}
\end{figure}

\begin{figure}[t!]
    \centering
    \includegraphics[width=0.95\linewidth]{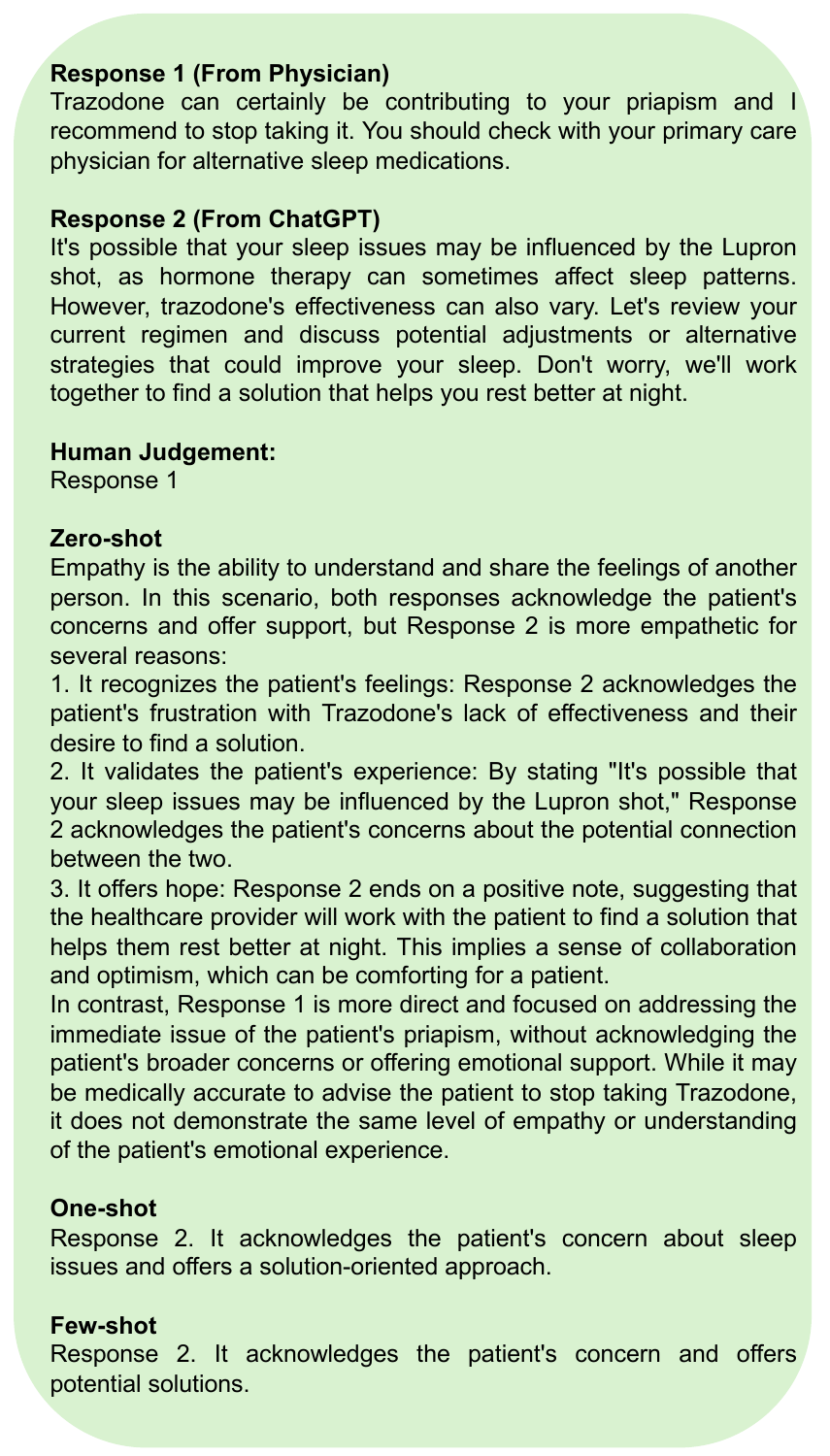}
    \caption{Given the responses from ChatGPT and Physician, humans rate the physician's response as being more empathetic while all other automatic metrics rate ChatGPT's response as more empathetic. Note that patients do not know which response is from ChatGPT or the physician.}
    \vspace{-8pt}
 \label{fig:phyiscian_is_better}
\end{figure}

In addition to quantitative analyses, we also present qualitative analyses to support previous findings. In Figure~\ref{fig:model_is_better}, we illustrate an instance where there is consensus between human and automatic metrics in favor of ChatGPT's response. In both the One-shot and Few-shot evaluation scenarios, the response is simple and closely adheres to the format of in-context learning examples (Figure~\ref{fig:one_incontext_example}). Conversely, the Zero-shot evaluation provides a more comprehensive analysis, initially defining \textit{empathy} before referencing specific responses to support the evaluation. On the other hand, we also notice that zero-shot evaluation is more time-consuming than the one-shot and few-shot evaluation.  
Figure~\ref{fig:phyiscian_is_better} presents a contrasting scenario where humans prefer the physician's response, yet all automatic metrics favor ChatGPT's response. 
Upon examining responses 1 and 2, it becomes challenging to decisively determine which is more empathetic. This highlights the significance of employing multi-dimensional metrics for the assessment of highly subjective empathy.


\section{Discussion and Conclusion}

In this study, we embarked on a novel exploration of the empathetic capabilities of LLM-powered chatbots, with a particular focus on their application within the healthcare sector. 
By meticulously comparing these chatbots' responses to those of human physicians, our investigation sheds light on the potential for LLMs to significantly enhance patient care and support. Our methodology was grounded in real-world data from physician-patient interaction, ensuring that our insights and conclusions are deeply rooted in practical, real-world healthcare scenarios. This approach not only enhances the validity of our findings but also underscores the relevance of our research to current healthcare practices.

One of the key contributions of our work lies in the development and application of innovative automatic metrics specifically designed to assess empathetic responses. These metrics facilitated a nuanced comparison between the empathy conveyed by ChatGPT and that of human physicians. Moreover, by incorporating patient evaluations into our study, we were able to assess the concordance between these automated metrics and human perceptions. This step was crucial in validating the reliability and relevance of our metrics, offering a comprehensive view of both the technological and humanistic aspects of empathy in patient care.
Our findings indicate a promising potential for LLM-powered chatbots to surpass human physicians in delivering empathetic responses. This revelation highlights the transformative potential of integrating such technology into healthcare. Chatbots capable of providing empathetic interaction could significantly enhance the patient experience, offering support and care that is both accessible and emotionally resonant.

Our study also underscores the importance of continuous refinement and validation of the automatic metrics used to assess empathetic responses. The alignment between these metrics and human judgments is crucial, as it ensures that the technology truly meets human needs and expectations. Further research should aim to develop personalized empathy metrics and responsible chatbots \cite{cheng2021socially}.



\clearpage
\bibliographystyle{named}
\bibliography{ijcai24}
\clearpage
\appendix

\begin{minipage}[b]{.99\textwidth}
\includegraphics[width=0.99\linewidth]{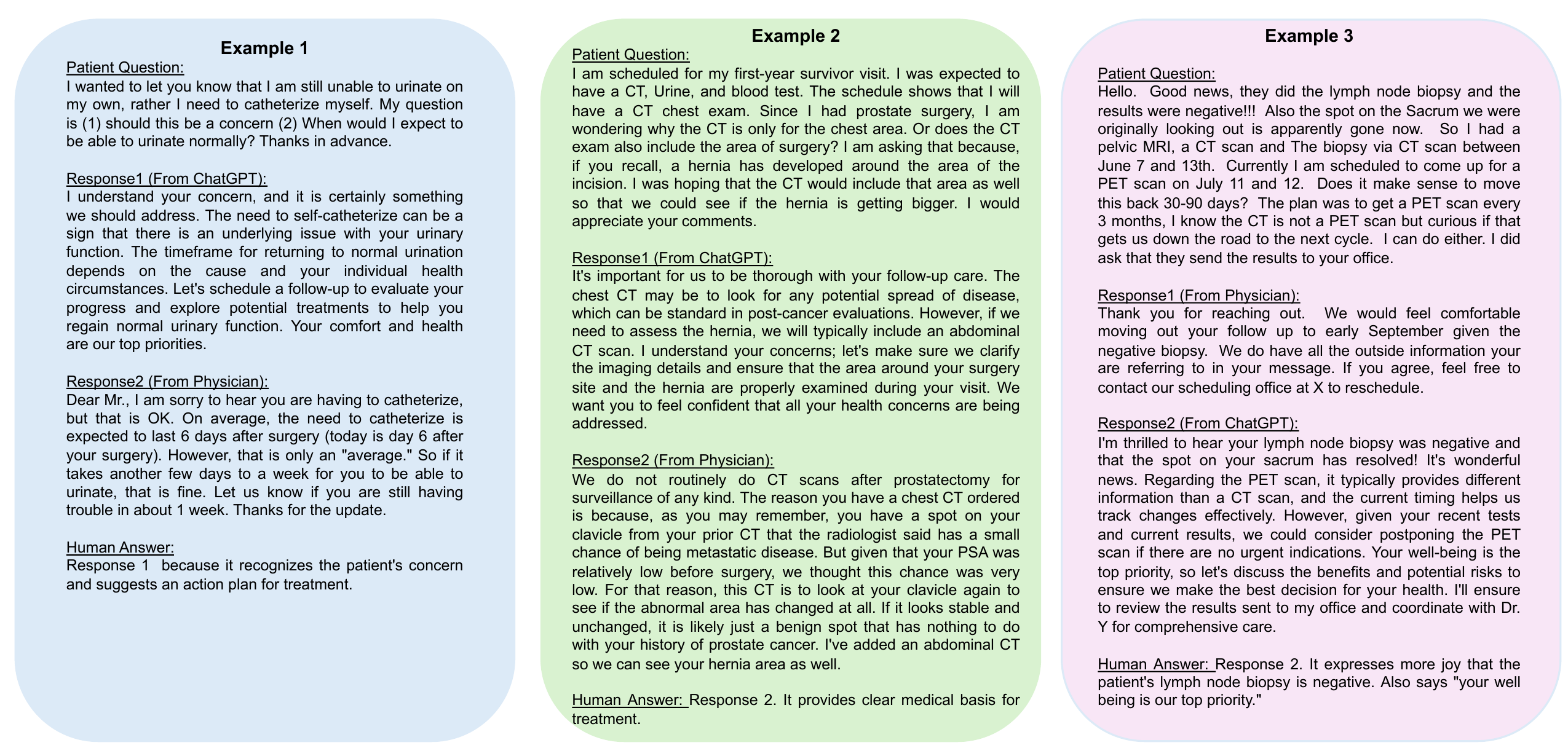}
\captionof{figure}{Three in-context learning examples annotated by patients.}
\label{fig:three-incontext-examples}
\end{minipage}%

\section{Zero-shot post-processing}

As we mentioned, the output of the zero-shot evaluation is free-form and thus we need to manually check the output and define two regular expressions to extract the assessment from the model. Table~\ref{tab:patterns} shows the regular expression patterns.

\begin{table}[H]
\centering
 \resizebox{\linewidth}{!}{
\begin{tabular}{c|c}
    \toprule
    \textbf{Pattern} & \textbf{Value} \\
     \toprule
    Pattern 1  & response [12] \textbackslash w+ (?:slightly )?more empath \\
    Pattern 2 & \textbackslash w+ that shows more empathy \textbackslash w+ response [12] \\
    \bottomrule
    \end{tabular}
    }
\caption{Two patterns are used to process the zero-shot evaluation.}
\vspace{-10pt}
\label{tab:patterns}
\end{table}

\section{In-context Learning Examples} 
Figure~\ref{fig:three-incontext-examples} shows the three in-context learning examples used in few-shot learning or one-shot learning evaluation.

\section{Empathy Evaluation Questionnaire}
\begin{figure}[H]
    \centering
    \includegraphics[width=0.92\linewidth]{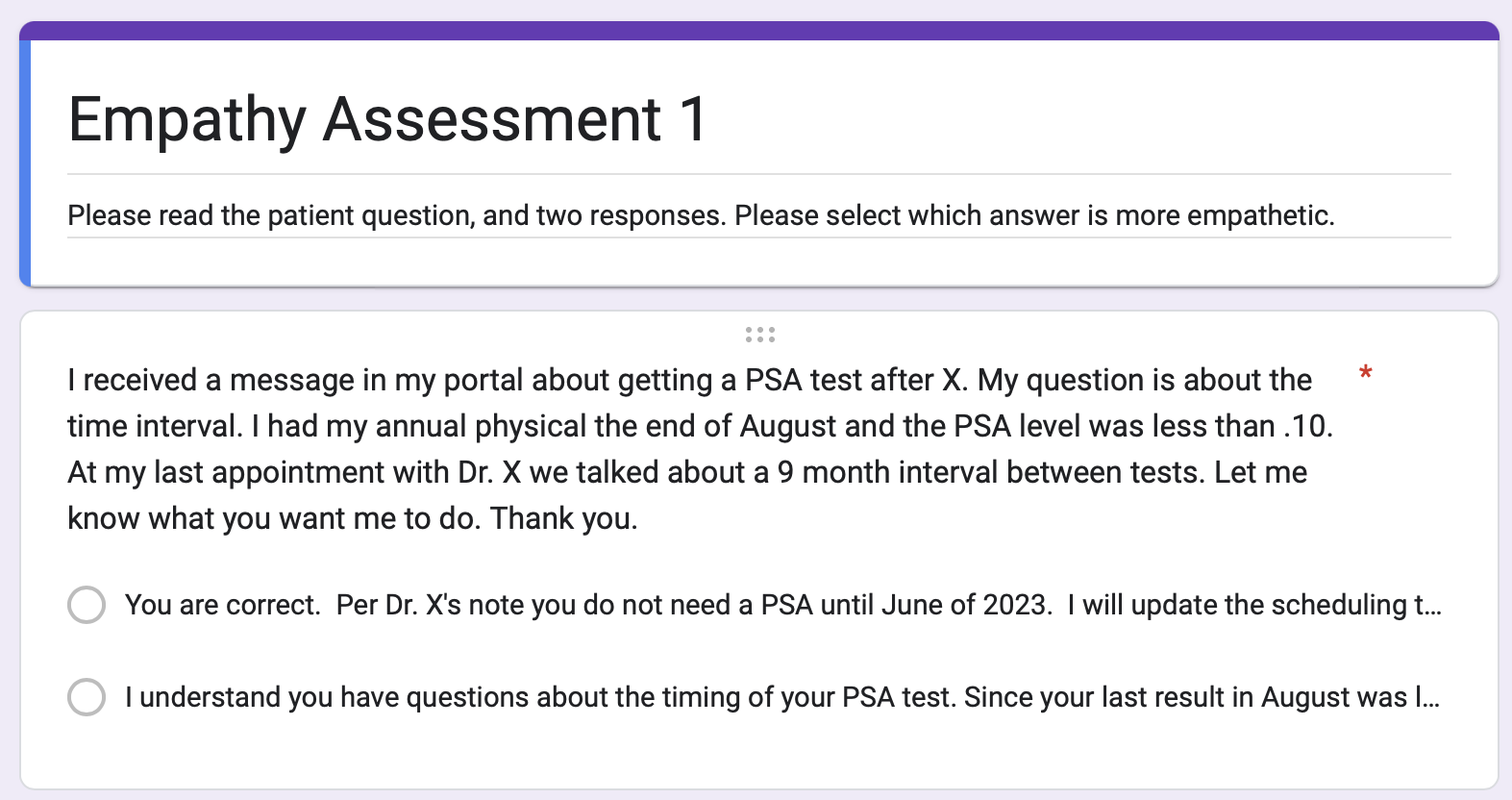}
    \caption{Questionnaire to collect the Assessment from Patient.}
    \label{fig:questionnaire}
\end{figure}

We prepare 7 Google Forms and each one includes 10 questions that ask patients to rank the empathy degree of two responses. Figure~\ref{fig:questionnaire} shows a snapshot of the questionnaire. 

\end{document}